\documentclass[conference]{IEEEtran}
\IEEEoverridecommandlockouts
\usepackage{cite}
\usepackage{multirow}
\usepackage{booktabs} 
\usepackage[table]{xcolor} 
\usepackage{amsthm}
\usepackage{amsmath,amssymb,amsfonts}
\usepackage{algorithmic}
\usepackage{graphicx}
\usepackage{textcomp}
\usepackage{xcolor}
\def\BibTeX{{\rm B\kern-.05em{\sc i\kern-.025em b}\kern-.08em
    T\kern-.1667em\lower.7ex\hbox{E}\kern-.125emX}}

\begin{document}

\title{Trajectory Prediction for Robot Navigation using Flow-Guided Markov Neural Operator}

\author{Rashmi Bhaskara*, Hrishikesh Viswanath*, and Aniket Bera\\
{Department of Computer Science, Purdue University, USA}\\
\thanks{*Equal Contribution}
\thanks{All authors are with the Department of Computer Science, Purdue University, USA,
        {\tt\small \{bhaskarr, hviswan, aniketbera\}@purdue.edu}}%
}


\maketitle

\begin{abstract}
Predicting pedestrian movements remains a complex and persistent challenge in robot navigation research. We must evaluate several factors to achieve accurate predictions, such as pedestrian interactions, the environment, crowd density, and social and cultural norms. Accurate prediction of pedestrian paths is vital for ensuring safe human-robot interaction, especially in robot navigation. Furthermore, this research has potential applications in autonomous vehicles, pedestrian tracking, and human-robot collaboration. Therefore, in this paper, we introduce \textbf{FlowMNO}, an Optical Flow-Integrated Markov Neural Operator designed to capture pedestrian behavior across diverse scenarios. Our paper models trajectory prediction as a Markovian process, where future pedestrian coordinates depend solely on the current state. This problem formulation eliminates the need to store previous states. We conducted experiments using standard benchmark datasets like ETH, HOTEL, ZARA1, ZARA2, UCY, and RGB-D pedestrian datasets. Our study demonstrates that FlowMNO outperforms some of the state-of-the-art deep learning methods like LSTM, GAN, and CNN-based approaches, by approximately 86.46\% when predicting pedestrian trajectories. Thus, we show that FlowMNO can seamlessly integrate into robot navigation systems, enhancing their ability to navigate crowded areas smoothly.
\end{abstract}


\section{Introduction}

Pedestrian trajectory prediction is an essential aspect of robotics research. It enables us to predict how pedestrians will move in various contexts. This feature is critical in robotics, particularly in guaranteeing safe interactions between robots and humans in complex and chaotic settings. Several applications make use of pedestrian trajectory prediction. In autonomous robot navigation \cite{li2020socially,morales2009autonomous,pradalier2005cycab,siagian2013mobile }, for example, self-driving cars employ a mechanism for predicting pedestrian trajectories to estimate future pedestrian coordinates and avoid collisions with pedestrians on the scene \cite{trivedi2018socially, carlone2015probabilistic, jain2018generative, kumar2019traphic, dosovitskiy2017end, dang2020bayesian}, thus ensuring safe navigation. Social robotics \cite{randhavane2016pedestrian,bera2019socially,bera2015adapt,bera2017realtime,bera2016glmp} enhances human-robot collaboration, particularly in crowded spaces, by using trajectory prediction to teach robots to avoid interrupting pedestrians or groups and maintain a safe distance, thereby avoiding any impact on their mental state. Robots tasked with crowd management at public events or transportation hubs benefit from trajectory prediction, as it optimizes traffic flow and enhances safety. 

\begin{figure}[!htb]
  \centering
  \includegraphics[width=0.9\columnwidth]{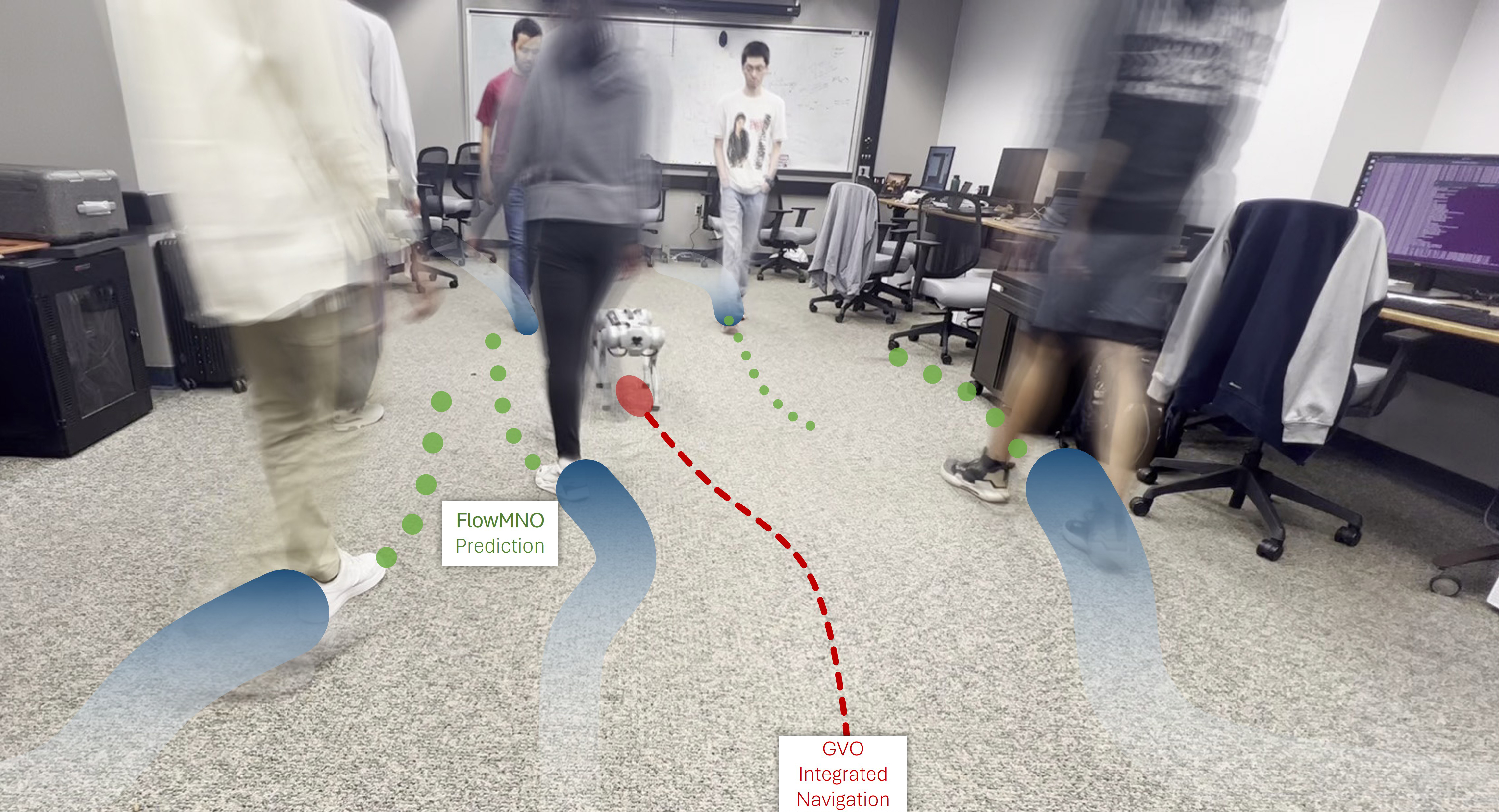}
  \caption{\textit{We present FlowMNO, an Optical Flow-Integrated Markov Neural Operator designed to capture pedestrian behavior across diverse scenarios. Our paper models trajectory prediction as a Markovian process, where future pedestrian coordinates depend solely on the current state. }}
  \label{fig:train}
\end{figure}

There are quite a few challenges in this domain. The complexity of human behavior, influenced by individual choices, social norms, and environmental factors, forms a formidable obstacle to achieving precise predictions. Furthermore, the inherent uncertainty in human movements presents a particularly demanding aspect of the problem. In crowded environments, the prediction of pedestrian trajectories becomes even more intricate due to the close proximity and interactions among individuals. Additionally, emergency situations necessitate rapid and accurate trajectory predictions, introducing an added layer of complexity as pedestrian behavior can drastically change under such circumstances. 

Numerous deep-learning models have been explored to predict the trajectory of objects in a scene. Convolutional Neural Network-based models are used to predict the trajectory from a sequence of frames \cite{yi2016pedestrian},\cite{yi2016pedestrian},\cite{varshneya2017human}. However, convolution filters cannot capture temporal relationships in the frames. Images are treated as independent entities. LSTMs are another family of neural network architectures that have been used for trajectory predictions \cite{alahi2016social},\cite{lee2017desire},\cite{huang2019stgat}. They are designed to capture temporal relationships. Some models have explored scaling the number of LSTM networks with the number of entities present in the scene \cite{bhaskara2023sg}. This would be computationally expensive. Furthermore, random chaotic motion cannot be modeled from temporal data. 

In our work, we model the trajectory of crowds as a Markov process, where the positions of the entities at a given time step only depend on the previous time step. The system is modeled using dissipative dynamics, where the model predicts the flow of the entities within the system. Rather than representing the system with discrete entities, we represent the scene with the \textit{flow} of the entity with respect to the previous time step, which is analogous to modeling the flow of a fluid. 

Neural Operators have shown promise in modeling various types of fluid behavior and dissipative chaotic systems \cite{li2021learning}. In this setting, we present a neural operator-based model \textbf{FlowMNO} to model the flow of entities, captured using optical flow estimation techniques. 

Our main contributions can be summarized as follows: 
\begin{enumerate}
  \item \textbf{Optical Flow Generation:} FlowMNO incorporates optical flow as a key component for input data generation. Optical flow is generated using the Farneback method in combination with a pedestrian detection algorithm. 
  \item \textbf{Modeling trajectories as a Markov process:} FlowMNO adopts a Markovian process model for pedestrian trajectory prediction. This model relies on the assumption that the future state of a pedestrian primarily depends only on their current state.
  \item \textbf{Comprehensive Evaluation:} The model's performance is evaluated on multiple datasets, including ETH and HOTEL \cite{pellegrini2009you}, ZARA1, ZARA2, UCY \cite{leal2014learning}, and RGB-D pedestrian \cite{Dataset-PedRGBD}. The evaluation includes a comparative study of FlowMNO against other deep-learning models commonly used in pedestrian trajectory prediction. The evaluation metric uses the average displacement error and final displacement error, providing a quantitative measure of prediction accuracy. We show that FlowMNO outperforms various state-of-the-art deep learning models by approximately 86.46\%.
\end{enumerate}

\section{Related Works}



In the realm of pedestrian trajectory prediction research, various deep learning models have been explored to address the intricate challenges inherent in this task. Among these models are \textbf{CNN-Based Approaches}, which encompass diverse techniques and exhibit unique technical characteristics. Noteworthy examples include Yi et al. \cite{yi2016pedestrian}, who introduced a deep neural network framework for understanding and predicting pedestrian behavior. Varshneya and Srinivasaraghavan \cite{varshneya2017human} proposed spatially aware deep attention models, enhancing spatial perception in human trajectory prediction. Yu et al. \cite{yu2020spatio} advanced the field with spatio-temporal graph transformer networks, adept at capturing complex pedestrian dynamics. Dan \cite{dan2020spatial} innovated by integrating a spatial-temporal block and LSTM network for trajectory forecasting, introducing temporal dependencies. Jain et al. \cite{jain2020discrete} presented a discrete residual flow model, adding probabilistic elements to pedestrian behavior prediction. Ridel et al. \cite{ridel2020scene} proposed a scene-compliant trajectory forecast model employing agent-centric spatio-temporal grids, enhancing predictive accuracy. Meanwhile, Zhang et al. \cite{zhang2021social} introduced the Social-IWSTCNN model, strategically incorporating social interactions into predictions. Zhao and Liu \cite{zhao2021stugcn} presented STUGCN, a social spatio-temporal unifying graph convolutional network, revolutionizing trajectory prediction approaches. Zamboni et al. \cite{zamboni2021pedestrian} focused on pedestrian trajectory prediction using convolutional neural networks. Mohamed et al. \cite{mohamed2020socialstgcnn} unveiled Social-STGCNN, an intricate social spatio-temporal graph convolutional neural network tailored for precise human trajectory prediction. It is worth noting that while these models exhibit impressive performance, they share common limitations, including challenges with generalization beyond their training data, computational intensity, and limited consideration of environmental factors and interactions with unpredictable agents. Ethical concerns, such as privacy and bias, also necessitate careful consideration in real-world deployments in diverse urban environments.

\textbf{LSTM-Based Models} have gained prominence for their ability to capture temporal dependencies and interactions in trajectory data. Alahi et al. presented the ``Social LSTM" method, focusing on predicting human trajectories in crowded spaces using LSTM networks \cite{alahi2016social}. SG-LSTM utilizes Social Group LSTM with group detection to enhance robot navigation in dense crowds, as proposed by R. Bhaskara et al. \cite{bhaskara2023sg}. Lee et al. introduced ``DESIRE," a framework for distant future prediction in dynamic scenes with interacting agents \cite{lee2017desire}. Fernando et al. proposed an LSTM framework with soft and hardwired attention mechanisms to predict trajectories and detect abnormal events \cite{fernando2018soft}. Moreover, works like ``Trajectron" \cite{ivanovic2019trajectron}, ``STGAT" \cite{huang2019stgat}, and ``Spatio-temporal Attention Model" \cite{haddad2019situation} have leveraged spatiotemporal modeling and attention mechanisms to enhance prediction accuracy. Monti et al. introduced ``Dag-net," a double attentive graph neural network for trajectory forecasting \cite{monti2020dag}. Finally, ``SSeg-LSTM" \cite{syed2019sseglstm} and ``Multi-agent Tensor Fusion" \cite{zhao2019multi} incorporate semantic scene segmentation and multi-agent fusion, respectively, to improve contextual trajectory prediction in diverse scenarios. Despite their advancements, these models often encounter challenges like accurate long-term trajectory forecasting, sensitivity to initial conditions, computational inefficiencies, handling multimodal predictions, limited context awareness, and interpretability issues. These challenges can impact their real-world applicability and require further research and development.

Recent advances in pedestrian trajectory prediction have witnessed the development of various \textbf{Generative Adversarial Networks (GANs) Based Models}. Gupta et al. introduced Social GAN, employing adversarial training to generate socially acceptable pedestrian trajectories \cite{gupta2018social}. Fernando et al. presented GD-GAN, which focuses on trajectory prediction and group detection in crowded scenes through GANs \cite{fernando2018gd}. Sadeghian et al. introduced SoPhie, an attentive GAN that predicts paths adhering to social and physical constraints \cite{sadeghian2019sophie}. Amirian et al. put forth Social Ways, utilizing GANs to model multi-modal distributions of pedestrian trajectories \cite{amirian2019social}. Kosaraju et al. developed Social-BiGAT, a multimodal trajectory forecasting model incorporating Bicycle-GAN and graph attention networks \cite{kosaraju2019socialbigat}. Lai et al. proposed Trajectory Prediction via Attended Ecology Embedding, integrating ecological embeddings into the GAN framework for trajectory prediction in heterogeneous environments \cite{lai2020trajectory}. Meanwhile, Huang et al. introduced STI-GAN, a multimodal pedestrian trajectory prediction model using spatiotemporal interactions within a GAN framework \cite{huang2021sti}. Despite their progress, these GAN-based models face challenges such as data dependency, subjective behavior definitions, adherence to physical constraints, handling contextual diversity, computational complexity, metric standardization, and generalization across diverse scenarios.

In conclusion, various trajectory prediction approaches, including CNN-based, LSTM-based, and GAN-based models, exhibit distinct limitations, ranging from generalization issues to computational complexities and ethical concerns. However, FlowMNO, as a Markovian process, addresses some of these challenges and provides more accurate long-term trajectory forecasts.

\begin{figure*}[!t]
    \centering
    \includegraphics[width=\textwidth]{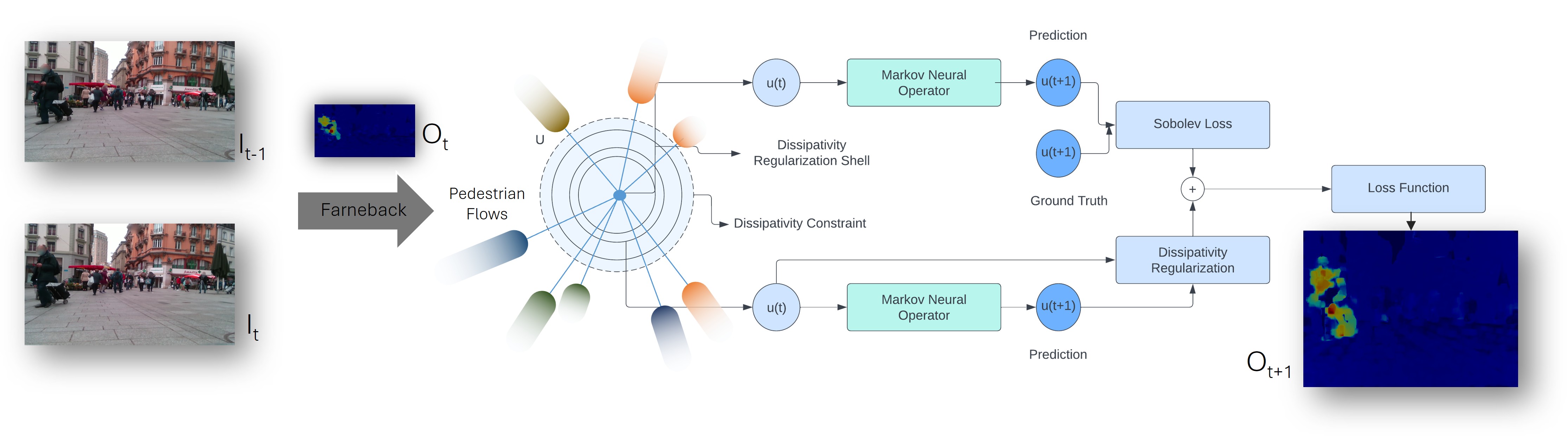}
    \caption{\textit{\textbf{Model Architecture}: FlowMNO incorporates optical flow as a key component for input data generation. Optical flow is generated using the Farneback method in combination with a pedestrian detection algorithm. FlowMNO adopts a Markovian process model for pedestrian trajectory prediction. This model relies on the assumption that the future state of a pedestrian primarily depends only on their current state.}}
    \label{fig:pipeline}
\end{figure*}

\section{Problem Formulation}

The problem of trajectory prediction of multiple moving entities within the environment observable to a mobile robot is modeled as a Markov process, where the positions of these entities are predicted solely based on their current positions. We make the following assumptions for the model
\textbf{Assumptions} 1. The observable scene has multiple entities. 2. The motions of the entities are independent of each other. 3. We only consider the motion of the entities within the frame of reference. 4. The overall motion of all entities within the scene is considered random. 5. The system as a whole is chaotic. 6. The system as a whole is modeled as a Markov process. 

\subsection{Modeling Randomness}

\textbf{Markov Process} A Markov Process, which is a random process, characterizes a sequence of events or states in which the probability of transitioning from one state to another depends solely on the current state, exhibiting the Markov property. This property implies that the future state of the process is conditionally independent of its past states, given the current state, making it memoryless. Markov Processes are extensively employed \cite{blitzstein2014introduction,puterman1994markov,sutton2018reinforcement} for modeling systems with inherent randomness and uncertainty, enabling the analysis and prediction of future states or events based on historical observations. The problem of trajectory prediction of a system of independent entities can be thought of as a Markov process because the system as a whole is too random to be governed by deterministic models. For instance, a person may suddenly change walking direction, and the previous states cannot be used to predict this phenomenon. 

\textbf{Transition Dynamics:} Probabilistically, the transition dynamics can be represented as:
\[
s_{t+1} \sim P(s_{t+1} | s_t, a_t)
\]
Here, \(a_t\) signifies the actions taken by an agent a, within the system, such as a robot or a person, at time step \(t\).

\textbf{Sobolev Norms and Solution Operators:} To model the solution operator, that learns to approximate the operator mapping the solution from the current to the next step \(\hat{S}_h: f(t) \mapsto f(t+h)\), we use Sobolev norm. The model approximates the solution operator \(S_h\), an element of the underlying continuous time semigroup \(\{S_t : t \in [0, \infty)\}\), using a neural operator. A neural operator, as presented in Li et al \cite{li2021learning}, is a neural network that can learn infinite dimensional functions. A neural operator has been shown to be powerful in modeling ergodic systems, dissipative dynamics, and chaotic systems. In this problem, we model the motion of crowds as a stochastic system, with groups of entities \textit{flowing} in and out of the observable frame of the robot. Crowds can be modeled on Riemann 3n Manifold in 3D Euclidean space \cite{ivancevic2010crowd}. However, from the point of view of a robot, entities get bigger as they move closer and smaller as they move away, denoting motion along the z-axis. They can move in a left-right direction. Movement along the up-down direction is minimal unless there are slopes in the environment. This entire system can be modeled as a dissipative stochastic process or an open thermodynamic system. If the inertial forces of an individual entity are weaker than the overall force of the system, the system can be thought of as being very crowded. Dissipative systems have a global attractor set or an invariant measure, that can be learned using the Sobolev norm. However, in a stochastic process, the attractor set is more random, representing a probability distribution of where the entities can converge. Human motion in most cases can be constrained by traffic rules, i.e., most people tend to walk along a sidewalk. This trajectory is always in the system. While there are infinitely many paths that people can take, they generally don't do so and move in a countably finite set of trajectories. 

Given the ground truth operator \(S_h\) and the learned operator \(\hat{S}_h\), along with the residual \(r = \hat{S}_h(f) - S_h(f)\), the neural operator computes the step-wise loss in the Sobolev norm as shown in \cite{li2021learning}, defined as:
\[
\|r\|_{k,p} = (\sum_{i=0}^{k} \|r^{(i)}\|_p^{p})^{1/p}
\]
Furthermore, for values of, \(p = 2\), the Sobolev norm can be efficiently computed in Fourier space as:
\[
\|r\|_{k,2}^2 = \sum_{n=-\infty}^{\infty} \left(1 + n^2 + \ldots + n^{2k}\right) |\hat{r}(n)|^2
\]
Here, \(\hat{r}\) represents the Fourier series of \(r\) and \(f\) represents the optical flow inputs. 

\textbf{Long-Term Predictions:} The neural operator exhibits the capability of making long-term predictions, leveraging the semigroup properties of the solution operator \(\hat{S}_h\). By repeatedly composing \(\hat{S}_h\) with its own output, long-time pedestrian trajectories can be approximated efficiently. Thus, for any \(n \in \mathbb{N}\), \(f(nh)\) is computed as follows:
\[
f(nh) \approx \hat{S}_h^n(f_0) := \underbrace{\hat{S}_h \circ \hat{S}_h \circ \ldots \circ \hat{S}_h}_{\text{n times}} (f_0)
\]

\textbf{Theoretical Foundation:} The neural operator is a solution operator that approximates the solution operator to a dynamic system that is locally Lipschitz \cite{li2021learning}. On a compact set $K$, the neural operator, denoted by, $\hat{S}_h$, estimates the system for any $n \epsilon \mathbf{N}$, within a specified error $\epsilon$, as  follows

\[
\sup_{f_0 \in K} \sup_{k \in \{1, \ldots, n\}} \lVert f(kh) - \hat{S}_h^k(f_0) \rVert_F < \epsilon.
\]

\section{FlowMNO Architecture}
\subsection{Optical Flow Estimation and Integration with MNO}

In FlowMNO, the input frame at time $t-1$ and time t, denoted as $I_{t-1}$ and $I_{t}$ respectively, are used to generate the optical flow using the Farneback Optical Flow estimation method. Optical flow estimation is a crucial component of our pedestrian trajectory analysis methodology. It serves as a fundamental tool for quantifying the motion of objects across consecutive frames in a video sequence. We adopt the Farneback method \cite{farneback2003two}, a widely used technique for optical flow computation. The process begins by initializing the YOLOv3-tiny model \cite{redmon2015yolo}, which is pre-trained with weight parameters and configurations to identify pedestrians within each frame. Subsequently, detected pedestrian bounding boxes, along with their associated confidences and class IDs, are extracted from the frames. To enhance the precision of pedestrian localization, we apply non-maximum suppression (NMS) based on confidence scores, ensuring that only the most reliable bounding boxes are retained.

The core of our optical flow calculation relies on the Farneback method, which is described by the following equations:

\[
I(x, y, t) \approx I(x + u, y + v, t + 1)
\]

where \(I(x, y, t)\) represents the intensity of a pixel at coordinates \((x, y)\) and time \(t\), and \((u, v)\) denotes the optical flow vector.

This method estimates the optical flow between the two frames $I_{t-1}, I_t$ and generates the optical flow $O_t$ at time step t for frame $I_{t}$. The optical flow results are visualized using a color map, and flow lines are drawn within the pedestrian bounding boxes to illustrate the motion patterns. Finally, the computed optical flow frames are saved for subsequent analysis or visualization. This iterative process is executed for each frame in the video sequence, enabling a comprehensive examination of pedestrian trajectories, their direction, and speed. This analysis provides valuable insights into crowd dynamics and behavior. 

The generated $O_t$ is then used as input to the Markov Neural Operator (MNO), which estimates the optical flow at time step $t+1$. The output generated by MNO is denoted as $O_{t+1}$. The FlowMNO pipeline as shown in Fig.\ref{fig:pipeline}, can be summarized as follows:

\begin{equation}
    O_t = \text{FarnebackOpticalFlow}(I_{t-1} \rightarrow I_t)
\end{equation}

\begin{equation}
    O_{t+1} = \text{MNO}(O_t)
\end{equation}

The optical flow information that MNO generates, denoted as $O_{t+1}$, is later utilized for estimating pedestrian trajectories from time step $t$ to $t+1$.

\begin{table*}[!t]
\caption{Comparison of Quantitative Results Using Error Metrics (ADE/FDE, lower is better for both) with State-of-the-Art Deep Learning Methods, Emphasizing FlowMNO's Performance}
\label{table:results}
\centering
\resizebox{0.9\textwidth}{!}{
\begin{tabular}{@{}ccccccc@{}}
\toprule
\textbf{Model} & \textbf{ETH} & \textbf{Hotel} & \textbf{Zara1} & \textbf{Zara2} & \textbf{UCY} & \textbf{Average (ADE/FDE)} \\
\midrule
SSeg-LSTM \cite{syed2019sseglstm} & 0.15 / 0.295 & 0.05 / 0.08 & 0.05 / 0.08 & 0.07 / 0.1 & 0.1 / 0.16 & 0.08 / 0.15 \\
SS-LSTM \cite{xue2018ss} & 0.2 / 0.37 & 0.08 / 0.13 & 0.08 / 0.11 & 0.07 / 0.12 & 0.2 / 0.24 & 0.13 / 0.19 \\
Scene-LSTM \cite{manh2018scene} & 0.18 / 0.34 & 0.25 / 0.29 & 0.37 / 0.33 & 0.19 / 0.1 & 0.25 / 0.03 & 0.21 / 0.20 \\
Social-LSTM \cite{alahi2016social} & 0.5 / 1.07 & 0.11 / 0.23 & 0.22 / 0.48 & 0.25 / 0.5 & 0.27 / 0.77 & 0.27 / 0.41 \\
Social-Attention \cite{vemula2018social} & 0.46 / 4.56 & 0.42 / 3.57 & 0.21 / 0.65 & 0.41 / 3.39 & 0.36 / 4.45 & 0.38 / 3.49 \\
Starnet \cite{zhu2019starnet} & 0.73 / 1.48 & 0.49 / 1.01 & 0.27 / 0.56 & 0.33 / 0.7 & 0.41 / 0.84 & 0.46 / 0.94 \\
SGAN(20V-20) \cite{gupta2018social} & 0.61 / 1.22 & 0.48 / 0.95 & 0.21 / 0.42 & 0.27 / 0.54 & 0.36 / 0.75 & 0.39 / 0.78 \\
CNN-based \cite{nikhil2018convolutional} & 1.04 / 2.07 & 0.59 / 1.17 & 0.43 / 0.90 & 0.34 / 0.75 & 0.57 / 1.21 & 0.59 / 1.22 \\
\rowcolor[rgb]{0.73, 0.85, 0.94} FlowMNO (Ours) & 0.04 / 0.02 & 0.02 / 0.01 & 0.05 / 0.09 & 0.01 / 0.02 & 0.02 / 0.04 & 0.03 / 0.04 \\
\bottomrule
\end{tabular}
}
\end{table*}

\subsection{Trajectory Estimation}

In the Trajectory Estimation stage of FlowMNO, we focus on estimating the trajectories of pedestrians from time step \(t\) to \(t+1\) using the optical flow information provided by the \(O_{t+1}\) frame. To achieve this, we leverage the centroid coordinates \(x\) and \(y\) of detected pedestrians within the bounding boxes in frame \(I_t\).

For each pedestrian in frame \(I_t\), we compute the displacement based on the optical flow information in \(O_{t+1}\). Specifically, we use the optical flow value at the centroid \((x, y)\) of the pedestrian's bounding box in frame \(I_t\), denoted as (dx,dy). This optical flow value indicates the displacement of the centroid from frame \(I_t\) to frame \(I_{t+1}\).

To estimate the trajectory for a pedestrian at time step \(t+1\), we calculate the new centroid coordinates \((x_{t+1}, y_{t+1})\) in frame \(I_{t+1}\) using the optical flow vectors, allowing us to estimate their trajectory for the next time step. This trajectory estimation is valuable for tracking pedestrians in dynamic scenarios and is a key component of FlowMNO's functionality.

\subsection{Integration of FlowMNO Trajectories with GVO Framework for Robot Navigation}

\begin{figure}[!htb]
  \centering
  \includegraphics[width=0.9\columnwidth]{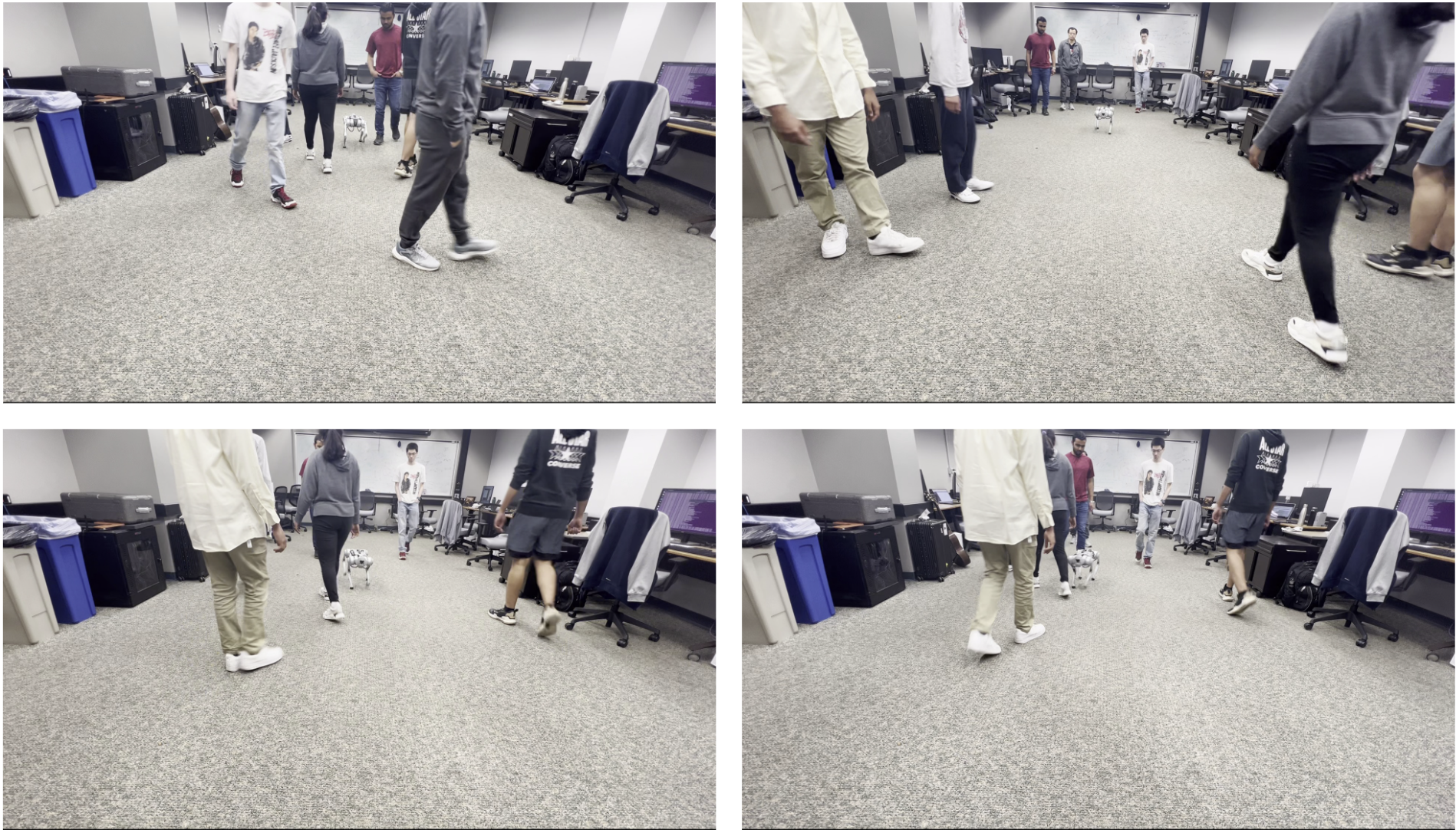}
  \caption{\textit{Demonstration of FlowMNO integrated with GVO, deployed on a Robot.}}
  \label{fig:train}
\end{figure}

In this integration, FlowMNO's predictions for pedestrians' future positions at time \(t+1\) (\(x_{t+1}\) and \(y_{t+1}\)) serve as dynamic obstacles within the Generalized Velocity Obstacles (GVO) framework, enhancing robot navigation in dynamic environments. GVO \cite{WilkieGVO}, a widely-used navigation technique, calculates velocity obstacles considering both robot and obstacle dynamics. FlowMNO, our innovative pedestrian trajectory prediction model, provides these predictions based on optical flow information, enabling the robot to anticipate pedestrian movements. The robot's constraints are defined by a state transition function \(R(t, u)\), where \(u\) represents control inputs, including steering angle (\(u_{\phi}\)) and speed (\(u_s\)), and \(t\) is time. Equation (\ref{eq:carpos}) characterizes the robot's position evolution.

\begin{equation}\label{eq:carpos}
    R(t, u) = \begin{pmatrix}
    \frac{1}{\tan(u_{\phi})}\sin(u_{s}\tan(u_{\phi})t) \\
    -\frac{1}{\tan(u_{\phi})}\cos(u_{s}\tan(u_{\phi})t) +  \frac{1}{\tan(u_{\phi})}
    \end{pmatrix}
\end{equation}

To incorporate the predicted pedestrian positions (\(x_{t+1}\) and \(y_{t+1}\)), we treat them as additional moving obstacles, modifying the GVO equations to account for their positions and velocities. The key to effective navigation within the GVO framework lies in calculating the relative velocity (\(v_{r_i}\)) between the robot and each pedestrian (\(i\)). This relative velocity is pivotal in assessing the potential collision risk and determining the robot's desired velocity (\(v_d\)) for collision avoidance. 

The relative velocity (\(v_{r_i}\)) between the robot and pedestrian \(i\) is computed using the following equation:
\begin{equation}
    v_{r_i} = v_{r_{\text{robot}}} - v_{r_i^{\text{pedestrian}}}
\end{equation}
Where \(v_{r_{\text{robot}}}\) represents the velocity of the robot relative to a stationary frame.
\(v_{r_i^{\text{pedestrian}}}\) represents the velocity of pedestrian \(i\) relative to the same stationary frame. For more information, please refer to \cite{WilkieGVO}.

By calculating \(v_{r_i}\) for each pedestrian \(i\), the GVO framework evaluates the dynamics of robot-pedestrian interactions. This information is then utilized to determine the robot's desired velocity (\(v_d\)), which ensures safe navigation while avoiding potential collisions. The desired velocity (\(v_d\)) is typically computed based on optimization criteria, considering factors such as minimum separation distance and collision avoidance strategies, and is an essential component of the robot's motion planning process within dynamic environments.

\section{Training}

\begin{figure}[!htb]
  \centering
  \includegraphics[width=0.9\columnwidth]{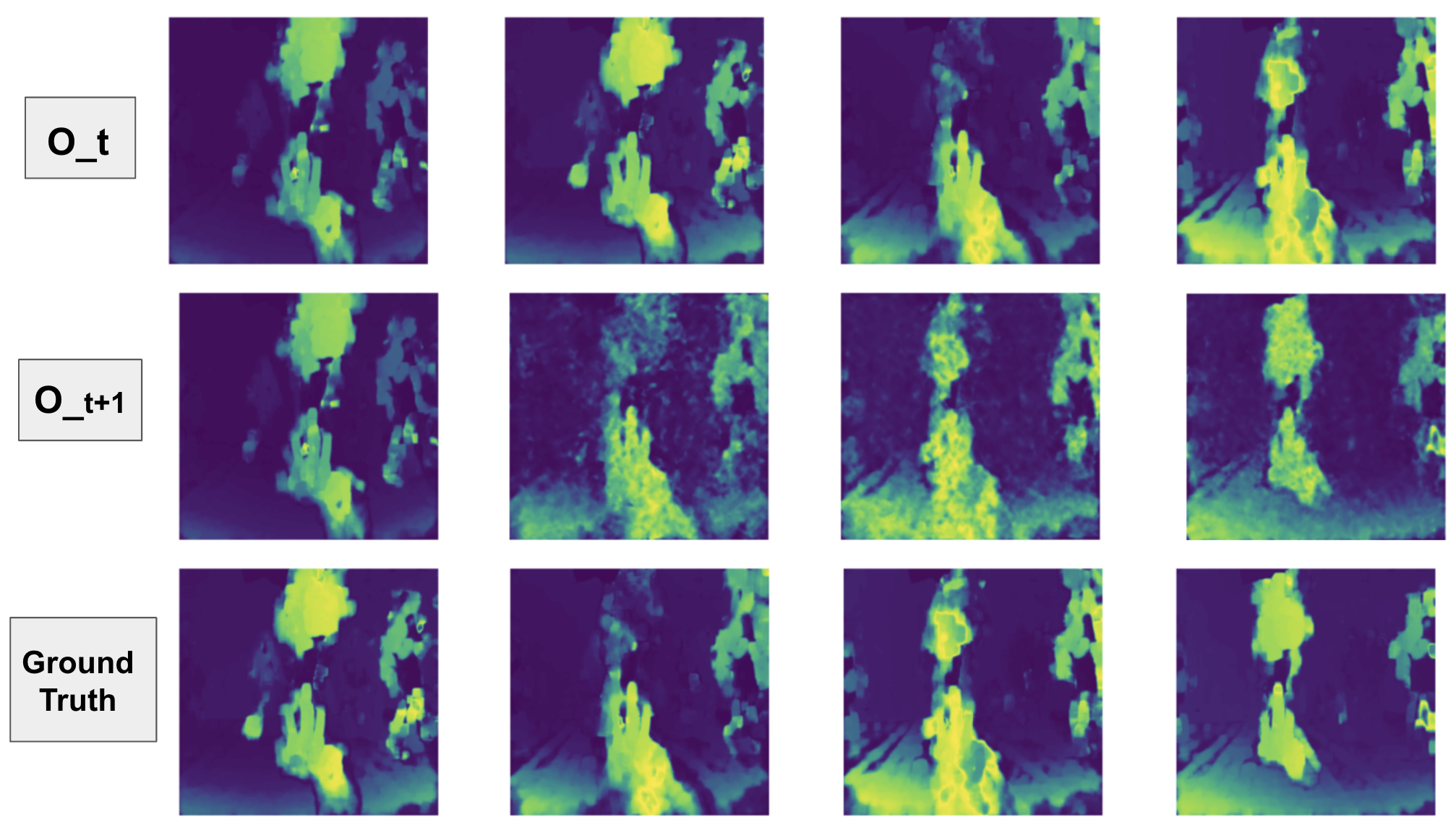}
  \caption{\textit{Training results: $O_t$ is the input optical flow frame at time step $t$, and $O_{t+1}$ is the output generated by the MNO model for time step $t+1$.}}
  \label{fig:train}
\end{figure}

FlowMNO underwent intensive training on the RGB-D pedestrian dataset \cite{Dataset-PedRGBD}, spanning 60 epochs. The dataset was split into 70$\%$ training, 20$\%$ validation, and 10$\%$ testing sets. A batch size of 5 was used for efficient training. Key parameters included a learning rate of 0.0005, a scheduler step size of 10, and a gamma value of 0.5. Training employed the Adam optimizer and a learning rate scheduler (step size: 10, gamma: 0.5) for stability. The loss function utilized during training was Mean Squared Error (MSE), a standard measure for regression tasks. The process harnessed the computational power of an Nvidia A30 GPU, significantly expediting performance optimization.

\section{Evaluation}

\subsection{Evaluation Metrics}

FlowMNO's performance is assessed using two key evaluation metrics:

\begin{enumerate}
    \item \textbf{Average Displacement Error (ADE)}: ADE measures the average Euclidean distance between the predicted and ground-truth pedestrian positions over multiple time steps. It quantifies the accuracy of FlowMNO's trajectory predictions by considering the entire prediction horizon.
    
    \item \textbf{Final Displacement Error (FDE)}: FDE quantifies the accuracy of FlowMNO's predictions at the final time step. It calculates the Euclidean distance between the predicted position at the last time step and the ground-truth position, providing insight into the model's ability to make accurate long-term predictions.
\end{enumerate}

These evaluation metrics offer a comprehensive assessment of FlowMNO's performance across various datasets and scenarios, enabling a quantitative analysis of its prediction accuracy.

\subsection{Results}

The evaluation results of FlowMNO on datasets such as ETH and HOTEL \cite{pellegrini2009you}, as well as ZARA1, ZARA2, and UCY \cite{leal2014learning}, using the established metrics, are presented in Table~\ref{table:results}. Our observations indicate that modeling pedestrian trajectory prediction as a Markovian process yields promising results. Additionally, we conducted experiments on the RGB-D pedestrian dataset \cite{Dataset-PedRGBD}, where we achieved notable performance with an ADE of 0.03 and an FDE of 0.04. These findings underscore the effectiveness of our approach in various real-world scenarios.

To further emphasize FlowMNO's performance, we calculated the reduction in Average Displacement Error (ADE) compared to other state-of-the-art models. FlowMNO significantly outperforms these models, achieving substantial reductions in ADE. The table \ref{table:ade-reduction} provides a clear comparison of FlowMNO's performance against several other state-of-the-art models in predicting pedestrian trajectories. The ``ADE Reduction (\%)" column indicates how much FlowMNO reduces the Average Displacement Error (ADE) compared to each model, with higher percentages representing better performance. On average, FlowMNO outperforms these models by approximately $86.46\%$. These significant improvements highlight FlowMNO's capability to provide highly accurate pedestrian trajectory predictions, making it a valuable tool for enhancing safety and efficiency in various trajectory prediction applications.

\begin{table}[ht]
\centering
\caption{ADE Reduction Comparisons}
\label{table:ade-reduction}
\resizebox{0.4\textwidth}{!}{
\begin{tabular}{|c|c|}
\hline
\textbf{FlowMNO vs. Model}                & \textbf{ADE {Reduction} (\%)} \\ \hline
SSeg-LSTM         & 50.00                       \\ \hline
SS-LSTM           & 69.23                       \\ \hline
Scene-LSTM        & 88.24                       \\ \hline
Social-LSTM       & 96.26                       \\ \hline
Social-Attention  & 98.88                       \\ \hline
Starnet           & 97.30                       \\ \hline
SGAN(20V-20)      & 96.72                       \\ \hline
CNN-based         & 98.06                       \\ \hline
\textbf{Average {Reduction}}    & \textbf{86.46}              \\ \hline
\end{tabular}
}
\end{table}

\section{Conclusion}

In this study, we introduced FlowMNO, a novel framework combining Markov Neural Operators (MNOs) and the Farneback optical flow estimation method for pedestrian trajectory prediction. FlowMNO predicts future pedestrian positions (t+1) based solely on the current time step (t), reducing the need for extensive historical data storage. This characteristic, coupled with FlowMNO's ability to model pedestrian movement as a chaotic system, holds great promise for real-time applications, such as autonomous navigation and crowd management.

Nevertheless, FlowMNO faces certain challenges. The computational demands of MNOs may pose obstacles for resource-constrained robots, demanding exploration of real-time optimization and hardware acceleration. Additionally, relying on optical flow estimation may introduce inaccuracies in scenarios with obstructions or complex scenes, impacting prediction accuracy. To improve FlowMNO's practicality, future research should focus on strategies like real-time optimization, sensor fusion to overcome optical flow limitations, model interpretability enhancements, data augmentation for generalization, and hybrid model integration to handle diverse scenarios effectively. Online learning mechanisms can further empower FlowMNO to adapt dynamically, solidifying its role in autonomous systems for enhanced safety and efficiency in urban environments. Thus, future research endeavors should focus on refining FlowMNO's robustness, adaptability, and real-time capabilities to fully harness its potential.

\end{document}